\pdfoutput=1

\documentclass[11pt]{article}

\usepackage[preprint]{acl}

\usepackage{times}
\usepackage{latexsym}

\usepackage[T1]{fontenc}

\usepackage[utf8]{inputenc}

\usepackage{microtype}

\usepackage{inconsolata}

\usepackage{graphicx}

\usepackage{algorithm}
\usepackage{algorithmic}

\usepackage{natbib}

\title{Reverse-Speech-Finder: A Neural Network Backtracking Architecture for Generating Alzheimer's Disease Speech Samples and Improving Diagnosis Performance\thanks{A patent application related to this work has been filed (United States Patent Application No.  63/810,191, pending).}}

\author{
Victor OK Li\thanks{Corresponding authors.}\thanks{Equal contributions.}, Yang Han\footnotemark[3], Jacqueline CK Lam\footnotemark[2]\footnotemark[3] \\ The University of Hong Kong \\ \{vli, yhan, jcklam\}@eee.hku.hk
\And
Lawrence YL Cheung \\ The Chinese University of Hong Kong \\ yllcheung@cuhk.edu.hk 
}

\begin{document}
\maketitle
\begin{abstract}
This study introduces Reverse-Speech-Finder (RSF), a groundbreaking neural network backtracking architecture designed to enhance Alzheimer's Disease (AD) diagnosis through speech analysis. Leveraging the power of pre-trained large language models, RSF identifies and utilizes the most probable AD-specific speech markers, addressing both the scarcity of real AD speech samples and the challenge of limited interpretability in existing models. RSF's unique approach consists of three core innovations: Firstly, it exploits the observation that speech markers most probable of predicting AD, defined as the most probable speech-markers (MPMs), must have the highest probability of activating those neurons (in the neural network) with the highest probability of predicting AD, defined as the most probable neurons (MPNs). Secondly, it utilizes a speech token representation at the input layer, allowing backtracking from MPNs to identify the most probable speech-tokens (MPTs) of AD. Lastly, it develops an innovative backtracking method to track backwards from the MPNs to the input layer, identifying the MPTs and the corresponding MPMs, and ingeniously uncovering novel speech markers for AD detection. Experimental results demonstrate RSF's superiority over traditional methods such as SHAP and Integrated Gradients, achieving a 3.5\% improvement in accuracy and a 3.2\% boost in F1-score. By generating speech data that encapsulates novel markers, RSF not only mitigates the limitations of real data scarcity but also significantly enhances the robustness and accuracy of AD diagnostic models. These findings underscore RSF's potential as a transformative tool in speech-based AD detection, offering new insights into AD-related linguistic deficits and paving the way for more effective non-invasive early intervention strategies.
\end{abstract}

\section{Introduction}
Alzheimer's disease (AD) is a progressive neurodegenerative disorder and the leading cause of dementia worldwide, affecting more than 55 million people around the world. The social and economic impact of AD is profound, with care and treatment costs exceeding hundreds of billions of dollars annually, placing a significant burden on families and healthcare systems \cite{alzfact}. Early and accurate diagnosis is crucial for timely intervention, which can slow disease progression and improve patient outcomes \cite{robinson2015dementia}.

Language impairment is one of the earliest signs of cognitive decline and can effectively discriminate AD patients from healthy individuals, providing a non-invasive and low-cost solution for large-scale AD screening \cite{mueller2018connected, petti2020systematic, richard2024linguistic}. However, the limited availability of AD speech samples poses significant challenges in the development of robust data-driven speech-based AD diagnostic models \cite{abrar2024survey, ding2024speech}.

To address these challenges, this study proposes Reverse-Speech-Finder (RSF), a novel domain-specific speech-based AD detection approach that leverages large language models (LLMs) for fine-grained speech analysis, including novel speech marker discovery and data generation based on such novel markers. RSF presents a novel reverse engineering AD detection approach to determine key speech markers predicting AD. Experimental results demonstrate that RSF has successfully uncovered novel AD speech markers, providing new insights into AD-related speech impairments. Moreover, by generating synthetic speech data enriched with these novel markers, RSF mitigates the scarcity of real AD speech samples, enhancing both model robustness and diagnostic accuracy. These findings underscore RSF's potential as a powerful tool for advancing speech-based AD detection and broadening our understanding of AD-related linguistic deficits.

\subsection{Alzheimer's Disease Diagnostic Markers}
AD can be characterized by a variety of biomarkers in different modalities, including demographic, genetic, cognitive, brain imaging, blood, and cerebrospinal fluid (CSF) markers \cite{elazab2024alzheimer}. Although CSF and brain imaging markers are the most reliable biomarkers for the diagnosis of AD, they are highly invasive or expensive \cite{lee2024robust}. In contrast, speech-based speech markers offer a low-cost and non-invasive alternative for early detection of AD that can be deployed at scale \cite{fraser2015linguistic, eyigoz2020linguistic}. Artificial intelligence (AI) approaches are pivotal in identifying key markers predicting AD, facilitating more accurate and robust diagnosis of AD \cite{li2021designing}. By leveraging AI techniques to extract critical speech markers indicative of AD, we can significantly enhance the precision of AD detection, support timely interventions, and deepen our understanding of the underlying mechanisms behind AD-related cognitive decline.

\subsection{Related Work and Research Gaps}
Research has increasingly focused on identifying speech markers that could predict the onset of AD using NLP techniques \cite{ding2024speech}. Early studies explored traditional machine learning models to analyze speech patterns and identify features correlated with AD. These studies mainly used hand-crafted features of speech, such as unique word count, speech length, and vocabulary diversity, combined with supervised learning algorithms such as support vector machines, random forests, and logistic regression \cite{petti2020systematic}. These traditional machine learning approaches have shown some promise in distinguishing between AD patients and healthy controls based on speech data, although they often struggle to capture the complexity of language impairments in AD \cite{shi2023speech}.

Recent advances in AD detection have been driven by the rapid development of deep learning techniques. In particular, transformer-based deep learning architectures have demonstrated the ability to automatically extract high-level features from raw speech data, improving the accuracy of AD classification \cite{balagopalan2020bert, ilias2023detecting, saltz2021dementia, valsaraj2021alzheimer}. However, two significant challenges hinder the clinical adoption of these deep learning models: limited training samples and lack of interpretability. Recent studies have sought to address these limitations by using LLMs to extract interpretable speech markers from speech data, and use these extracted markers to predict AD with improved accuracy and interpretability \cite{mo2024leveraging, heitz2024linguistic}. Moreover, by leveraging the text generation capabilities of LLMs, the speech markers have been used to generate new speech samples for the training of diagnostic models and to improve predictive performance \cite{mo2025dect}. These LLM-driven approaches aim to identify key speech characteristics linked to AD by capitalizing on the prior knowledge encoded in LLMs pre-trained on massive amounts of data, making them more suitable for clinical applications.

Despite recent advances in speech-based AD detection using deep learning and NLP techniques, several key gaps remain to be addressed. Firstly, existing AD speech studies focus only on associative speech markers and do not prioritize the most probable speech markers that best capture the specific characteristics of speech and language most indicative of AD. These most probable markers could include novel speech markers with stronger diagnostic and prognostic relevance for AD. This oversight limits the ability of existing speech-based diagnostic models to detect early and critical signs of AD. Secondly, limited speech data continue to hinder the generalization of speech-based diagnostic models. Effective data generation and augmentation methods to increase the sample size of speech samples to capture the most probable speech markers of AD have not been fully explored.

\subsection{Research Significance and Objectives}
This study presents a significant advance in speech-based AD research by introducing RSF, a pioneering interpretable reverse engineering methodology designed to identify novel speech markers that can play a crucial role in the etiology of AD. The innovative approach proposed in this research offers substantial improvements over existing marker-finding studies focused on identifying speech markers for AD, addressing critical challenges in interpreting and discovering the most probable markers. This study aims to harness the immense potential of LLMs by using speech data from individuals diagnosed with AD, as well as normal control (NC), to fine-tune a pre-trained LLM. This model is explicitly optimized for classifying AD, allowing the extraction of meaningful insights from complex speech data. The findings of this study underscore the importance of innovative methodologies in advancing our understanding of complex diseases such as AD. Our proposed reverse engineering approach reveals the novel most probable speech markers of AD, potentially advancing the landscape of speech-based AD research. The objectives of this study include the following.
\begin{enumerate}
    \item Develop the RSF architecture to identify the most probable AD speech markers.
    \item Investigate different speech tokens and the speech markers they represent at the word and category level in the RSF architecture.
    \item Compare the performance of AD detection task using (a) raw speech samples, (b) speech generated via speech markers, and (c) speech generated via novel most probable AD speech markers identified by RSF.
\end{enumerate}

RSF is significantly different from prior work. While Reverse-Gene-Finder (RGF) \cite{li2025unravelling} focuses on identifying causal genetic markers of AD from single-cell gene expression data using a pre-trained genomic foundation model, our work investigates the most probable speech markers indicative of AD using a pre-trained language model. This makes it possible to enhance AD diagnostic performance using speech data generated from these most probable AD speech markers. Moreover, our approach differs from Causal Tracing \cite{meng2022locating}, which primarily aims to identify the causal neurons responsible for specific language model predictions. In contrast, our method traces neuron activations backward to uncover the most probable AD speech markers. Further, unlike the data generation method proposed by \cite{mo2024leveraging}, which focuses on augmenting training data through known AD speech markers, our method generates new speech samples based on the most probable AD speech markers identified via RSF.

 The significance of RSF lies not only in generating diverse and representative speech data to improve AD detection but also in identifying novel speech markers that can be used for AD diagnosis and prognosis, offering significant advantages in advancing speech-based AD studies. The insights gained from RSF will deepen our understanding of AD speech decline and hold significant potential for application in different disease contexts, offering a versatile framework to uncover the most probable factors in a wide range of classification tasks beyond medical diagnostics.

\section{Research Methodology}
\begin{figure}[t]
  \centering
  \includegraphics[width=\columnwidth]{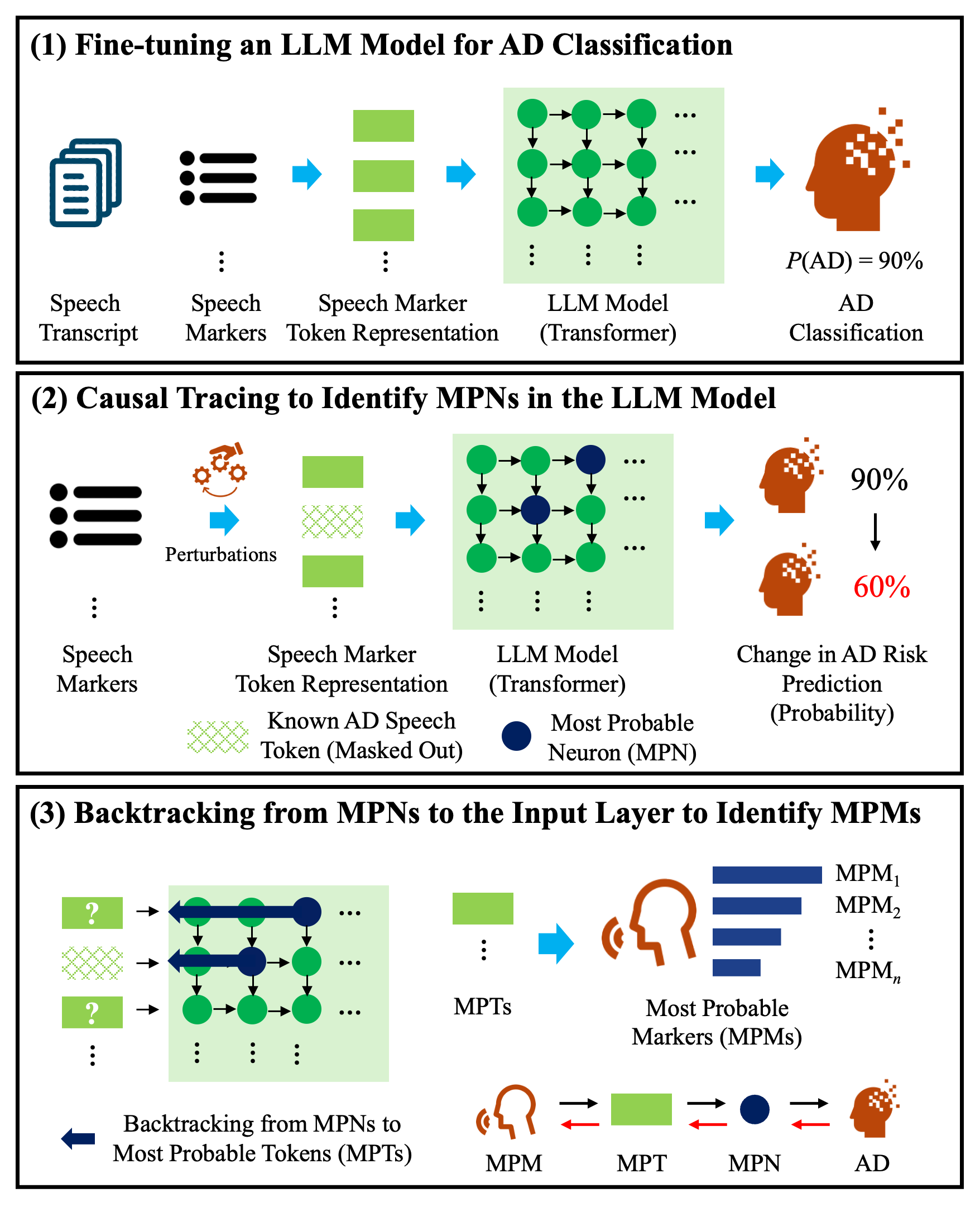}
  \caption{Overview of Reverse-Speech-Finder (RSF)}
  \label{fig:RSF}
\end{figure}

RSF proceeds in three stages (see Figure \ref{fig:RSF}). Firstly, we fine-tune an LLM specifically for AD classification, utilizing available speech data from AD and NC subjects. Secondly, by modifying the input data by masking out known speech markers strongly associated with AD, we systematically identify the most probable neurons (MPNs) related to AD, employing causal tracing techniques to observe the effects of perturbations of such known markers on neurons across different layers of the fine-tuned LLM. Finally, RSF backtracking enables the identification of the most probable tokens (MPTs) at the input layer, and the most probable markers (MPMs) they represent, most likely to activate MPNs.

\subsection{RSF and Its Novelties}
\subsubsection{Discovery of MPMs and MPNs}
We exploit the observation that speech markers with the highest probability of predicting AD, defined as the MPMs, must have the highest probability of activating those neurons (in the neural network model) with the highest probability of predicting AD, defined as the MPNs.

How can we locate the MPNs with the highest probability of predicting AD? This study adopts the causal tracing technique originally developed for LLM-based text generation \cite{meng2022locating} to identify the MPNs in an LLM for speech-based AD classification with the following assumptions and modifications. We assume that when key information about the AD-associated speech markers (obtained from previous AD speech marker studies) in the input data is masked (or perturbed), the probability of AD prediction will likely change, making it possible to observe the effects of in-silico perturbations of known AD speech markers on disease onset prediction, which are mediated by neurons in the LLM.

Based on this assumption, the difference between the probability of getting AD in (1) the baseline corrupted run (where AD-associated speech markers are corrupted by noise) and (2) the corrupted-with-restoration run (where AD-associated speech markers remain corrupted, but a particular neuron (i.e., hidden state) is reverted to the clean version) is calculated to measure the indirect effect of the hidden state in the model. The indirect effects of hidden states across different layers are then used to quantify the importance of different hidden states and select the MPNs indicative of AD.

\subsubsection{Introduction of Speech Token Representation at the Input Layer}
We employ a speech token representation at the input layer, where speech samples are encoded as speech tokens. By backtracking from the MPNs to the input layer, we identify the MPTs that are highly indicative of AD. These MPTs, along with their corresponding MPMs, enable the generation of new synthetic speech data that capture speech impairments that are potentially overlooked in the original dataset. The newly generated data can then be used to train and improve the performance of speech-based AD detection models.

In this study, we focus on speech markers extracted from speech transcripts at two levels: word and category. These markers provide insight into various aspects of speech deficiencies, making them valuable indicators for AD detection. Specifically, each speech sample consists of a set of speech markers carrying different speech characteristics, expressed as a sequence of words (e.g., ``cookie'', ``jar'', etc.) or categories (e.g., ``word-finding difficulties'', ``reduced syntactic complexity'', etc.). These speech markers are tokenized as a vector of speech tokens: $\{x_1, x_2, ..., x_T\}$, where each token is obtained from a pre-trained language model (e.g., GPT-2). For example, consider the speech marker $X=\textrm{``reduced syntactic complexity''}$ at the category level. When processed through a pre-trained language model, it may be tokenized into: $\{T_{1}=\textrm{``reduced''}, T_{2}=\textrm{``syntactic''}, T_{3}=\textrm{``complexity''}\}$. After backtracking, the identified MPTs may include: $\{T_{1}, T_{2}, T_{3}\}$. These MPTs are then mapped uniquely back to the original speech marker $X$. Similarly, at the word level, each word can be further decomposed into one or more tokens.

The speech token representation in the input layer allows each speech marker (known or novel to AD) to be represented as a discrete and unique entity in the input space. Some speech tokens correspond to well-established AD speech markers (as specified in the model input), while others, previously unknown, can be discovered through the backtracking process. This token representation enables the identification of new speech markers predictive of AD, improving the accuracy and interpretability of speech-based AD detection.

\subsubsection{Development of Backtracking Method}
In contrast to the existing neural network architectures, which track neuron activations from the input layer to the output layer in a feed-forward manner, we develop an innovative backtracking method to track backwards from the MPNs to the input layer, identifying the MPTs.

The information flowing from the input layer to the MPNs is backtracked and summarized to identify the MPTs more likely to activate these MPNs. Assume that the neural network model has $L$ layers. The $i$th hidden state at layer $l$ is denoted as $h_i^l$. The identified MPNs are denoted as $\{\hat{h}_i^l\}$. The activation (i.e., indirect causal effect) of each MPN $\hat{h}_i^l$ is denoted as $\textrm{IE}(\hat{h}_i^l)$. For each input, the score of a speech token at position $i$ is quantified by the weighted sum of the accumulated indirect effect of all MPNs backtracked to the $i$th unit in the input layer. Specifically, the score of the ith unit at layer $l$, denoted as $s_i^l$, is a weighted sum of all interconnected MPNs in the next layer $l+1$, where each MPN $k$ is quantified by the indirect effect $\textrm{IE}(\hat{h}_k^{l+1})$ and the score $s_k^{l+1}$ at layer $l+1$. This score is computed iteratively from the last layer to the input layer. The weights W measure the strength of interconnectivity between one hidden state and another hidden state across layers. It can be determined by the fitted model parameters, e.g., self-attention weights in transformers. Speech token scores are calculated and averaged after iterating every position in each input sample (see Algorithm \ref{alg:RSF} in Appendix \ref{sec:appendix_backtracking} for more details).

The speech tokens with the highest scores are identified as MPTs. Once the MPTs are identified, the corresponding MPMs can be identified. Each MPM is assigned a score based on the sum of its contributing tokens. These MPMs, which capture the key speech characteristics of AD, can be used to generate novel speech data to improve speech-based AD diagnosis.

\subsection{Performance Comparison of RSF with Other LLM-enhanced Speech Generation Methodologies}
How can we show that the novel speech markers identified by RSF are the most probable ones? To demonstrate that the novel speech markers identified by RSF are the most probable ones, we conducted a series of experiments comparing the effectiveness of RSF-identified markers with those identified by other LLM-enhanced speech marker identification methods. These experiments aim to evaluate whether the markers identified by RSF significantly improve speech-based AD detection.

In particular, we compare the performance of AD diagnostic models trained on several types of speech data: (1) the original raw speech data, (2) speech data generated using AD speech markers identified by traditional interpretability methods, and (3) speech data generated from speech markers identified by RSF. By comparing these models, we aim to assess whether the speech markers discovered by RSF are more effective in capturing the most probable AD-specific speech characteristics, leading to better model performance.

This performance comparison aims to show that RSF-identified markers can improve AD detection accuracy and offer more reliable, clinically relevant indicators of cognitive decline. This will help establish the RSF approach as a promising method for identifying the most probable speech markers of AD, ultimately improving speech-based diagnostic tools for early detection and intervention of AD.

\section{Experimental Setup}
\subsection{Dataset}
We used the Pitt English Corpus dataset\footnote{\url{https://dementia.talkbank.org/access/English/Pitt.html}} \cite{becker1994natural}, a prominent AD audio speech dataset for AD research, derived from DementiaBank \cite{lanzi2023dementiabank} for our speech-based AD study. The Pitt dataset focuses on spontaneous connected speech, making it a valuable resource for benchmarking AD detection methods and advancing our understanding of the speech features associated with cognitive declines among AD patients. During the Pitt study, participants were asked to perform a picture description task using the Cookie Theft picture, a cognitive-linguistic assessment proposed by the Boston Diagnostic Aphasia Examination (BDAE) \cite{giles1996performance}. Spontaneous connected speech samples were collected to capture their linguistic abilities and cognitive functions.

\subsection{Data Preprocessing}
We selected the subjects labeled NC and AD in the Pitt dataset. Some participants were followed up by multiple clinical visits and had more than one speech sample. Longitudinal speech samples were considered cross-sectional speech samples. Our experiment included 390 speech transcript samples: 212 ADs and 178 NCs. Speech samples were pre-processed in three steps.

Firstly, we used a speech transcription model, OpenAI's Whisper model \cite{radford2023robust}, to convert audio samples into transcripts (i.e., text samples). We used speaker diarization \cite{bredin2023pyannote} to remove the examiner's speech.

Secondly, we leveraged the information extraction capabilities of the latest GPT-4o model to extract speech markers at the category level (e.g., ``word-finding difficulties'') using OpenAI's Chat Completion API \cite{openai_api}, following the methodology developed by \citet{mo2024leveraging, mo2025dect}. A fixed temperature of 1 was used for text generation. Specifically, the following prompting strategy was used for the extraction of speech markers at the category level: \textit{Identify linguistic patterns, keywords, or phrases that could potentially indicate cognitive impairment or Alzheimer's disease.}

Finally, we used a stratified 80/10/10 split of all speech samples to obtain the training, validation, and testing sets while preserving the distribution of diagnosis labels in each set. The training set was used for model development, the validation set for model selection, and the testing set for model evaluation. The training set included both original and new speech samples generated by speech markers (either from known markers or from RSF). The validation and testing sets were held out and only included the original speech samples, i.e., the speech-based AD diagnostic performance evaluation was based on the real speech samples.

\subsection{Identification of MPMs by RSF} \label{subsection:MPM_identification}
For our proposed RSF approach, we (1) fine-tuned an open-source pre-trained BERT and GPT-2 models \cite{radford2019language, devlin2019bert} for AD classification based on speech markers, (2) located the MPNs related to AD in the fine-tuned BERT or GPT-2 model, and (3) identified the MPTs (and the corresponding MPMs) using backtracking.

We selected the known AD-associated speech markers from previous AD speech studies. The speech markers at the word level included: ``apron'',  ``asking'',  ``boy'',  ``cabinet'',  ``cookie'', ``counter'', ``cupboard'', ``curtains'', ``dishcloth'', ``dishes'', ``drying'', ``exterior'', ``falling'', ``faucet'', ``floor'', ``girl'', ``jar'', ``kitchen'', ``mother'', ``overflowing'', ``plate'', ``sink'', ``stealing'', ``stool'', ``washing'', ``water'', ``window'', and ``woman'' \cite{eyigoz2020linguistic}. The speech markers at the category level included: ``word-finding difficulties'', ``semantic paraphrase'', ``circumlocutions'', ``reduced syntactic complexity'', and ``topic maintenance issues'' \cite{mo2025dect}.

By masking out known AD-associated speech markers, we adopted the causal tracing technique developed by \citet{meng2022locating} to identify the MPNs. The number of samples per input was set to 10 (i.e., one original sample plus nine corrupted samples). The noise level (the variance of the Gaussian noise distribution) was set to 1. The maximum sequence length (i.e., the maximum number of speech tokens) was set to 512 to reduce computational costs when iterating over all input data. Any input speech token vector exceeding this length was truncated. After locating the MPNs, we used backtracking to identify the MPTs and their corresponding MPMs.

\subsection{Diagnostic Model Training}
For diagnostic model development and performance comparison, the same pre-trained BERT or GPT-2 model was used as the backbone transformer model for our proposed RSF approach and the baseline models throughout our experiments. A classification layer was added to the pre-trained model for binary classification (i.e., NC or AD). Our proposed RSF approach and the two baselines are listed below.

\textbf{Speech generated via MPMs identified by RSF (our proposed approach)}:
We first used RSF to identify the most probable AD speech markers (i.e., MPMs), and each speech marker was assigned a weight based on its score. We then used these speech markers identified by RSF to generate new speech samples utilizing the data generation methodology developed by \citet{mo2025dect}. Specifically, we generated new speech transcript samples with the same diagnosis label in a 1:1 ratio, based on the latest GPT-4o model via OpenAI's Chat Completion API. The following prompt was used for speech transcript generation based on the speech markers at the word/category level: \textit{Here is a subject's speech transcript: [transcript]. Based on the transcript, generate a new speech transcript with the following words / linguistic features: [randomly selected speech markers]}. In addition, we used weighted randomization to select speech markers so that the most probable speech markers with higher scores were more likely to be involved in the speech data generation process. After speech data generation via RSF-identified markers, we fine-tuned the model for AD classification using the raw speech samples plus the newly generated speech samples from the training set.

\textbf{Raw speech samples}: We fine-tuned the model for AD classification using the raw speech transcript samples from the training set.

\textbf{Speech generated via IG- or SHAP-based speech markers}: We fine-tuned the model for AD classification using the raw speech transcript samples plus the newly generated speech samples from the training set. The newly generated speech samples were obtained using the same speech data generation methodology as detailed in the proposed approach above, except that it did not use RSF-based markers but used the speech markers identified via Integrated Gradients (IG) \cite{sundararajan2017axiomatic} or SHAP analysis \cite{lundberg2017unified}.

\subsection{Evaluation Metrics}
We selected the best AD classification model using the validation set and evaluated the fine-tuned model on the testing set. This was repeated five times, and we reported the average performance metrics. We used accuracy and F1-score for model performance evaluation. They are two commonly used evaluation metrics in classification tasks. Accuracy, ranging from 0 to 1 (the higher, the better), measures the percentage of correctly detected AD cases. F1-score, also ranging from 0 to 1 (the higher, the better), combines the precision (positive predictive value) and recall (sensitivity) scores and provides a more comprehensive evaluation of detection accuracy.
 
\subsection{Experimental Settings}
 We fine-tuned the pre-trained GPT-2 (about 137M parameters) and BERT (about 110M parameters) models with a learning rate of 1e-4 and a weight decay of 0.01 after performing trials with a range of hyperparameter values (learning rate of 1e-3, 1e-4, and 1e-5, and weight decay of 0.1 and 0.01). The training process consisted of 20 epochs with a batch size of 16. The best proposed and baseline models were selected based on the accuracy of the AD classification task on the validation set. The computing infrastructures are detailed in Appendix \ref{sec:appendix_computing_infrastructures}.

\section{Results and Discussion}
\subsection{Performance Comparision}
Table~\ref{tab:performance_comparison} presents the average accuracy and F1-score across five runs for Alzheimer's disease (AD) detection using two base models (BERT and GPT-2) and four different configurations: (1) raw speech, (2) speech generated using IG-based markers, (3) speech generated via SHAP-based markers, and (4) speech generated using our proposed RSF-based markers. Across both models, F1-scores consistently exceed accuracy scores, suggesting that the classifiers are better at identifying true AD cases than non-AD (NC) cases. This pattern highlights a bias toward higher sensitivity, which is often desirable in medical screening tasks.

RSF-based speech generation yields the best performance across both models. For GPT-2, RSF achieves the highest accuracy (85.6\%) and F1-score (86.9\%), outperforming all baseline methods. In particular, RSF significantly outperforms the best baseline (SHAP with GPT-2) by 3.5\% in accuracy and 3.2\% in F1-score, according to the Wilcoxon signed-rank test (\textit{p} = 0.03).

These results demonstrate the robustness of the RSF framework across different language models and underscore its effectiveness in enhancing synthetic speech data for improved AD classification. By automatically identifying the most probable AD markers, RSF supports more targeted and informative data generation, leading to better generalization and model performance.

\begin{table*}
  \centering
  \scalebox{0.75}{
   \begin{tabular}{llcc}
      \hline
      \textbf{Model} & \textbf{Configuration} & \textbf{Avg. Accuracy (\%)} & \textbf{Avg. F1 (\%)} \\
      \hline
      BERT & Raw speech & 77.8 & 79.4 \\
      BERT & IG-based speech generation & 78.6 & 80.6 \\
      BERT & SHAP-based speech generation & 79.5 & 80.0 \\
      BERT & RSF-based speech generation & 82.1 & 82.6 \\
      \hline
      GPT-2 & Raw speech & 78.2 & 80.2 \\
      GPT-2 & IG-based speech generation & 78.6 & 79.7 \\
      GPT-2 & SHAP-based speech generation & 82.1 & 83.7 \\
      GPT-2 & RSF-based speech generation & \textbf{85.6} & \textbf{86.9}\\
      \hline
    \end{tabular}
  }
  \caption{Performance Comparison Across Different Data Generation Methods and Models at the Word Level}
  \label{tab:performance_comparison}
\end{table*}

\subsection{Key Speech Markers Identified by RSF}
In addition to performance improvement, our proposed RSF approach offers new insights into the linguistic mechanisms underlying AD by identifying the MPTs and the corresponding MPMs.

Table \ref{tab:speech_markers_word_level} in the Appendix summarizes the key speech markers at the word level. Specifically, based on
the causal tracing and backtracking analysis, the top MPTs were identified by RSF and their corresponding MPMs included: ``dishes"", ``water'', ``cookie'', ``cookies'', ``washing'', ``see'', ``getting'', ``trying'', ``running'', and ``mm'' (see Table \ref{tab:speech_markers_word_level}).

The comparison between the associative speech markers identified by SHAP analysis using the same diagnostic model (as listed in Table \ref{tab:speech_markers_word_level_SHAP} in the Appendix) and those identified by RSF (as listed in Table \ref{tab:speech_markers_word_level}) highlights few overlaps but significant differences, providing valuable information to improve AD detection. These differences suggest that RSF and SHAP capture distinct speech markers, with RSF uncovering novel speech markers that may be particularly indicative of AD-related language impairments. One key overlap is the speech marker ``mm'', which appears in both RSF and SHAP results. This suggests that filler sounds or hesitation markers may be strong indicators of AD-related speech impairments. Moreover, RSF has identified speech markers that differ from those found via SHAP, emphasizing words related to daily activities and actions, such as ``dishes'', ``water'', ``cookie'', ``washing'', and ``running''. These words are likely linked to narrative recall and semantic fluency impairments, common in AD patients, particularly when describing everyday scenes like the Cookie Theft Picture Description Task. Additionally, RSF highlights action verbs, such as ``getting'' and ``trying'', which may indicate difficulties in verb usage and syntactic construction in AD-related speech deterioration.

Table \ref{tab:speech_markers_category_level} in Appendix \ref{sec:appendix_speech_markers_category_level} summarizes the key speech markers at the category level (see Appendix \ref{sec:appendix_speech_markers_category_level} for more details). Most of them were initially not known as AD-associated speech markers (as listed in Section \ref{subsection:MPM_identification}). Although both sets of speech markers include ``word-finding difficulties'', underscoring its significance, RSF expands the scope by identifying additional discourse-related impairments such as ``irrelevant and unrelated information'' and ``inconsistent use of pronouns''. Moreover, the initial known speech markers focus more on syntactic and semantic aspects, including ``semantic paraphrase'', ``circumlocutions'', and ``reduced syntactic complexity'', whereas RSF captures a broader range of disruptions, including ``fragmented sentences'', ``repetition of words/phrases'', and ``vague or ambiguous references''. This shift suggests that RSF not only refines known speech markers, but also enhances interpretability by incorporating more granular speech characteristics. Furthermore, the presence of markers, such as ``frequent pauses and filler words'' and ``topic shifts'', provides additional insights into AD-related speech and language deficiencies.

\section{Discussion and Conclusion}
This study introduces RSF, an innovative approach aimed at improving speech-based diagnostic models for AD. The primary aim of RSF is to identify and utilize the most probable speech markers indicative of AD, addressing two significant challenges: the scarcity of AD speech samples and the limited interpretability of speech-based diagnostic models.

The study leverages the capabilities of LLMs to perform fine-grained speech analysis, including the discovery of novel speech markers and the generation of synthetic speech samples based on these markers. The RSF methodology presents several novel contributions to the field: (1) Backtracking Architecture: RSF introduces a unique neural network backtracking architecture that identifies key speech markers most indicative of AD. It focuses on speech markers with the highest probability of activating neurons associated with AD predictions, termed as MPNs. (2) Speech Token Representation: RSF employs a speech token representation at the input layer, enabling the identification of MPTs and their corresponding MPMs. (3) Improved Data Generation: By generating synthetic speech data enriched with novel markers, RSF addresses the scarcity of real AD speech samples, enhancing model robustness and diagnostic accuracy.

Experimental results demonstrate that RSF significantly improves the performance of AD detection models, outperforming traditional interpretability methods like SHAP and Integrated Gradients by 3.5\% in accuracy and 3.2\% in F1-score. The RSF approach not only enhances diagnostic accuracy but also provides deeper insights into the linguistic mechanisms underlying AD, identifying novel speech markers linked to narrative recall and semantic fluency impairments, as well as syntactic construction difficulties.

The current study can be further improved. Future work can incorporate additional speech samples from the DementiaBank database, including the TAUKADIAL Challenge dataset \cite{luz2024connected}, which consists of 222 mild cognitive impairment (MCI) and 164 NC samples from both English- and Chinese-speaking populations. This expansion will allow us to further assess the generalizability of our approach across linguistic and demographic variations. Moreover, future work can use open-source models, such as LLaMA and DeepSeek, to ensure better transparency in LLM-based data generation.

The RSF framework represents a significant advancement in speech-based AD research, offering a more refined and targeted approach to identifying speech markers most indicative of AD. By addressing critical challenges in interpreting and discovering these markers, RSF holds substantial potential for improving early detection and intervention strategies for AD.

The insights gained from RSF deepen our understanding of AD-related linguistic deficits and underscore the importance of innovative methodologies in advancing complex disease research. While the current study demonstrates promising improvements, future work should focus on expanding speech datasets to further strengthen speech-based AD diagnostic models and incorporating diverse linguistic and demographic variations to ensure broader applicability.

Overall, RSF contributes to the landscape of speech-based diagnostics by providing a versatile framework to uncover the most probable factors in a range of classification tasks beyond medical diagnostics, ultimately improving the accuracy and interpretability of speech-based AD detection.

\section*{Limitations}
In this study, RSF demonstrates promising improvements in AD speech-based diagnostics by identifying the MPMs and generating new speech data based on these identified MPMs. One key limitation is that data generation cannot fully replace real-world data collection. Although RSF effectively generates synthetic speech samples enriched with MPMs, these samples are still approximations based on the patterns learned from existing datasets and pre-trained models. The accuracy and reliability of MPM identification depend on the quality and diversity of real AD speech data, which means that a more extensive data collection is essential to refine and validate these identified markers. Increasing the availability of real AD speech samples from diverse populations and linguistic backgrounds will improve the robustness of RSF, ensuring that the identified markers generalize across different demographics. Future research should focus on expanding AD speech datasets to further strengthen speech-based AD diagnostic models.

\section*{Ethics Statement}
One key consideration in the effectiveness of RSF is the potential risks due to bias in model training. Since RSF relies on pre-trained LLMs, the speech markers it identifies are inherently shaped by the data these models were trained on. If the training data lack diversity in terms of demographics, language variations, or disease stages, RSF may identify markers that do not generalize well across all populations, leading to disparities in diagnostic accuracy.

\section*{Acknowledgments}
ChatGPT-4o was used to improve the language of the
manuscript and to assist in the writing of the manuscript (only covering literature review and discussion of results). The authors reviewed the content generated and
took full responsibility for the content of the manuscript.

This work was supported in part by the United States National Academy of Medicine Healthy Longevity Catalyst Award (Grant No. HLCA/E-705/24), administered by the Research Grants Council of Hong Kong, awarded to V.O.K.L. and J.C.K.L, and by The Hong Kong University Seed Funding for Collaborative Research 2023 (Grant No. 109000447), awarded to V.O.K.L. and J.C.K.L.

\bibliography{custom}

\newpage
 
\appendix

\section{RSF Backtracking}\label{sec:appendix_backtracking}
Algorithm \ref{alg:RSF} details the backtracking process in the RSF architecture.

\begin{algorithm}[htb]
\caption{RSF Backtracking Method}
\label{alg:RSF}
\textbf{Input}: Input Position $i$, MPNs  $\{\hat{h}_i^l\}$, Model Weights $W$\\
\textbf{Parameter}: Number of Layers $L$\\
\textbf{Output}: Speech Token Score at Position $i$

\begin{algorithmic}[1] 
\STATE Initialize $s_i^L=0$
\FOR{$l$ \textbf{from} $L-1$ \textbf{to} $1$}
    \STATE $s_i^l=\sum_{k} W_{i,k}^l * (\textrm{IE}(\hat{h}_k^{l+1}) + s_k^{l+1})$
\ENDFOR
\STATE \textbf{return} $s_i^1$
\end{algorithmic}
\end{algorithm}

\section{Computing Infrastructures}\label{sec:appendix_computing_infrastructures}
All experiments were carried out using a Nvidia A100 40GB GPU on a Linux system through Google Colab, with Python (version 3.11.11), and deep learning packages, including PyTorch (version 2.5.1+cu124) and Transformers (version 4.48.3). The total computational budget was approximately 3 GPU hours. The pre-trained 12-layer GPT-2 model was obtained from HuggingFace (MIT license)\footnote{\url{https://huggingface.co/openai-community/gpt2}}. The pre-trained BERT model was obtained from HuggingFace (Apache 2.0 license)\footnote{\url{https://huggingface.co/google-bert/bert-base-uncased}}. The causal tracing analysis was developed based on the code on GitHub (MIT license)\footnote{\url{https://github.com/kmeng01/rome/}}.

\section{Key Speech Markers Identified by RSF at the Category Level}
\label{sec:appendix_speech_markers_category_level}

For speech markers at the category level, based on the causal tracing and backtracking analysis, the top MPTs identified by RSF included: `word', `difficulties', `sentences', `errors', `repetition', `fragmented', `logical', `lack', `phrases', and `ition'. Speech tokens, including `word', `sentences', and `phrases', highlight the structural aspects of speech, while `difficulties', `errors', and `lack' indicate cognitive-linguistic struggles. The presence of `repetition' and `fragmented' suggests that patients with AD may exhibit frequent repetition of words/phrases and incomplete thoughts, while the inclusion of `logical' suggests that AD patients often struggle to maintain logical progression in spontaneous speech. The token `ition` by itself is not a standalone word, but is commonly seen as a suffix of longer words in the context of AD, such as `repetition', `transition', `cognition', etc.

\begin{table}
  \centering
  \scalebox{0.75}{
      \begin{tabular}{lcc}
        \hline
        \textbf{Speech Marker (Word)} & \textbf{RSF Score} \\
        \hline
        dishes & 1.57 \\
        water & 1.17 \\
        cookie & 0.91 \\
        cookies & 0.78 \\
        washing & 0.78 \\
        see & 0.65 \\
        getting & 0.65  \\
        trying & 0.52 \\
        running & 0.52 \\
        mm & 0.52 \\\hline
      \end{tabular}
      }
  \caption{The Most Probable AD Speech Markers Identified by RSF at the Word Level}
  \label{tab:speech_markers_word_level}
\end{table}

\begin{table}
  \centering
  \scalebox{0.75}{
      \begin{tabular}{lcc}
        \hline
        \textbf{Speech Marker (Word)} & \textbf{SHAP Score} \\
        \hline
        mm & 2.88 \\
        breakfast & 2.82 \\
        action & 2.76 \\
        huh & 2.35 \\
        garden & 1.88 \\
        tilting & 1.69 \\
        tumble & 1.49  \\
        alright & 1.48 \\
        counterspace & 1.43 \\
        motioning & 1.43 \\\hline
      \end{tabular}
      }
  \caption{The Most Associated AD Speech Markers Identified by SHAP Analysis at the Word Level}
  \label{tab:speech_markers_word_level_SHAP}
\end{table}

\begin{table}
  \centering
  \scalebox{0.8}{
      \begin{tabular}{lcc}
        \hline
        \textbf{Speech Marker (Category)}  & \textbf{RSF Score} \\
        \hline
        Repetition of words/phrases & 0.41 \\
        Fragmented sentences & 0.36 \\
        Word-finding difficulties & 0.32 \\
        Incomplete sentences & 0.28 \\
        Difficulty maintaining coherent narrative & 0.26 \\
        Frequent pauses and filler words & 0.22 \\
        Irrelevant and unrelated information & 0.20  \\
        Inconsistent use of pronouns & 0.19 \\
        Vague or ambiguous references & 0.18 \\
        Topic shifts & 0.18 \\\hline
      \end{tabular}
      }
  \caption{The Most Probable AD Speech Markers Identified by RSF at the Category Level}
  \label{tab:speech_markers_category_level}
\end{table}

To derive the corresponding MPMs from these tokens, we grouped related MPTs into meaningful speech markers and assigned each marker a score based on the sum of its contributing tokens. However, many of these speech markers exhibited strong similarities, e.g., ``repetition of words or phrases'' and ``repetition of phrases or words". To address this redundancy, we applied K-means clustering to further group similar speech markers based on their text embeddings capturing their semantic meanings \cite{reimers-2019-sentence-bert}. Within each cluster, the speech marker closest to the centroid in the embedding space was selected as the representative marker, ensuring a more concise and interpretable set of speech markers for further interpretation.

\end{document}